\documentclass[conference]{IEEEtran}
\IEEEoverridecommandlockouts
\usepackage{cite}
\usepackage{amsmath,amssymb,amsfonts}
\usepackage{algorithmic}
\usepackage{graphicx}
\usepackage{textcomp}
\usepackage{xcolor}

\usepackage{algorithm}
\usepackage{algorithmic}

\usepackage[listings]{tcolorbox}
\tcbuselibrary{listings,theorems}
\definecolor{fig1_yellow}{HTML}{fff4cc}
\definecolor{fig1_blue}{HTML}{dae8fc}
\definecolor{fig1_green}{HTML}{d5e8d4}
\definecolor{fig2_pink}{HTML}{ffebf8}

\definecolor{llm_blue}{HTML}{4A90E2}
\definecolor{llm_orange}{HTML}{F5A623}
\definecolor{llm_green}{HTML}{7ED321}
\definecolor{llm_red}{HTML}{FF6F61}

\definecolor{plap_yellow}{HTML}{fff2cc}
\definecolor{plap_purple}{HTML}{e1d5e7}
\definecolor{plap_blue}{HTML}{dae8fc}
\definecolor{plap_red}{HTML}{f8cecc}
\definecolor{plap_green}{HTML}{d5e8d4}

\definecolor{skill_plan_1}{HTML}{E6E6E6}
\definecolor{skill_plan_red}{HTML}{cc0000}
\definecolor{skill_plan_green}{HTML}{00cc00}
\definecolor{skill_plan_blue}{HTML}{0000cc}

\definecolor{purple}{HTML}{8d3a94}
\newtcbtheorem[]{exmp}{Prompt}%
{colback=purple!5,colframe=purple!80,fonttitle=\bfseries, left=.02in, right=.02in,bottom=.02in, top=.02in}{exmp}

\usepackage[
colorlinks=true,
linkcolor=blue,
urlcolor=blue,
citecolor=blue,
]{hyperref}

\usepackage{booktabs}
\usepackage{url}

\def\BibTeX{{\rm B\kern-.05em{\sc i\kern-.025em b}\kern-.08em
    T\kern-.1667em\lower.7ex\hbox{E}\kern-.125emX}}
\DeclareRobustCommand*{\IEEEauthorrefmark}[1]{%
    \raisebox{0pt}[0pt][0pt]{\textsuperscript{\footnotesize\ensuremath{#1}}}}
\begin{document}

\title{Empowering LLMs with Parameterized Skills for Adversarial Long-Horizon Planning}


\author{
    \IEEEauthorblockN{
        Sijia Cui\IEEEauthorrefmark{1,2,*},
        Shuai Xu\IEEEauthorrefmark{3,4,5,*},
        Aiyao He\IEEEauthorrefmark{1,3,4},
        Yanna Wang\IEEEauthorrefmark{1},
        Bo Xu\IEEEauthorrefmark{1,3}
    }
    \IEEEauthorblockA{\IEEEauthorrefmark{1}Institute of Automation, Chinese Academy of Sciences, Beijing, China}
    \IEEEauthorblockA{\IEEEauthorrefmark{2}School of Artificial Intelligence, University of Chinese Academy of Sciences, Beijing, China}
    \IEEEauthorblockA{\IEEEauthorrefmark{3}Nanjing Artificial Intelligence Research of IA, Nanjing, China}
    \IEEEauthorblockA{\IEEEauthorrefmark{4}University of Chinese Academy of Sciences,Nanjing, Nanjing, China}
    \IEEEauthorblockA{\IEEEauthorrefmark{5}Nanjing University of Information Science \& Technology, Nanjing, China}
    \IEEEauthorblockA{\IEEEauthorrefmark{*}Equal Contribution; Corresponding Author: Bo Xu\textless{boxu@ia.ac.cn}\textgreater.}
}

\maketitle

\begin{abstract}
Recent advancements in Large Language Models(LLMs) have led to the development of LLM-based AI agents. A key challenge is the creation of agents that can effectively ground themselves in complex, adversarial long-horizon environments. Existing methods mainly focus on (1) using LLMs as policies to interact with the environment through generating low-level feasible actions, and (2) utilizing LLMs to generate high-level tasks or language guides to stimulate action generation. However, the former struggles to generate reliable actions, while the latter relies heavily on expert experience to translate high-level tasks into specific action sequences. To address these challenges, we introduce the Plan with Language, Act with Parameter (PLAP) planning framework that facilitates the grounding of LLM-based agents in long-horizon environments. The PLAP method comprises three key components: (1) a skill library containing environment-specific parameterized skills, (2) a skill planner powered by LLMs, and (3) a skill executor converting the parameterized skills into executable action sequences. We implement PLAP in MicroRTS, a long-horizon real-time strategy game that provides an unfamiliar and challenging environment for LLMs. The experimental results demonstrate the effectiveness of PLAP. In particular, GPT-4o-driven PLAP in a zero-shot setting outperforms 80\% of baseline agents, and Qwen2-72B-driven PLAP, with carefully crafted few-shot examples, surpasses the top-tier scripted agent, CoacAI. Additionally, we design comprehensive evaluation metrics and test 6 closed-source and 2 open-source LLMs within the PLAP framework, ultimately releasing an LLM leaderboard ranking long-horizon skill planning ability. Our code is available at \url{https://github.com/AI-Research-TeamX/PLAP}.
\end{abstract}


\section{Introduction}
Planning in long horizons has always been a challenge, particularly in adversarial domains~\cite{ontanon2018first, ma2023large, nayak2024longhorizon}. It requires maintaining consistency in long-horizon decision-making~\cite{pirk2020modeling} (ensuring coherence between successive decisions) and adaptability (perceiving and responding to changes in the environment and opponent’s behavior)~\cite{ma2023large}.
Classical approaches~\cite{krishnan2017ddco,fox2017multi,akbari2016task,amiranashvili2018motion,wan2024lotus,zhao2023learning,konidaris2018skills,ames2018learning} for tackling long-horizon tasks include expert-driven rule-based systems, heuristic-based methods, hierarchical options learning, and skill learning. 
However, these methods often require extensive domain knowledge, incur extra training costs, and struggle to generalize across domains. 
Moreover, most of these are primarily designed for long-term motion and manipulation tasks, limiting their applicability in more complex, strategic, and adversarial settings.

\begin{figure}[t]
    \centering
    \includegraphics[width=\linewidth]{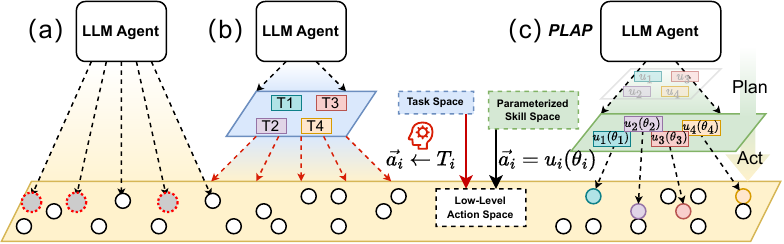}
    \caption{LLM agents Grounding. (a) shows the simplest grounding method, constraining the LLM agent output space to the environment's low-level \colorbox{fig1_yellow}{action space}. 
    (b) abstracts the space to high-level \colorbox{fig1_blue}{task space}, leaving the alignment between high-level and low-level intractable. 
    There exists an intermediate level, \colorbox{fig1_green}{parameterized skill space}, mapping to executable actions using predefined skill functions in (c).}
    \label{fig:action_space}
\end{figure}

Recently, Large Language Models (LLMs) trained on an enormous quantity of textual datasets, have demonstrated remarkable capabilities in various fields~\cite{devlin-etal-2019-bert, rae2021scaling, touvron2023llama, NEURIPS2020_1457c0d6, achiam2023gpt, hubinger2024sleeper, denison2024sycophancy, team2023gemini, reid2024gemini}. 
As the advancement of emergent reasoning and planning abilities, increasing LLM-based agents emerge for handling various generic tasks~\cite{brooks2024large, zhao2024expel, imani-etal-2023-mathprompter, hazra2024saycanpay, ding2024large, tang-etal-2024-medagents}.
One approach focuses on grounding LLM agents in the task environment by directly prompting the internal language model to generate low-level actions~\cite{pmlr-v162-huang22a, 10341989, ichter2022do, hazra2024saycanpay}, as illustrated in Fig.~\ref{fig:action_space}(a). 
Another approach demands the output action space of agents to high-level tasks, and ultimately translates the sequential planning of these high-level tasks into corresponding low-level actions, as shown in Fig.~\ref{fig:action_space}(b). 
Both methods have inherent limitations~\cite{Prakash_Oates_Mohsenin_2024, pan2024hi, ijcai2024p627, dalal2024planseqlearn, ma2023large}: the former struggles with invalid actions and the curse of dimensionality, while the latter faces challenges in effectively bridging high-level planning with execution.

Inspired by recent advancements in robotics~\cite{kumar2024practice}, we propose PLAP, a framework that grounds LLM agents using parameterized skills for adversarial long-horizon planning tasks. As depicted in Fig.~\ref{fig:action_space}(c), by leveraging predefined parameterized skill functions as an intermediary layer, PLAP mitigates the difficulties of coordinating different levels of abstraction.
\underline{\textbf{PLAP}} operates in two stages: 
1. \underline{\textbf{P}}lan with \underline{\textbf{L}}anguage: dynamically constructs prompt to guide the LLM in planning within the parameterized skill space;
2. \underline{\textbf{A}}ct with \underline{\textbf{P}}arameters: executes skill functions with corresponding parameters to generate a queue of low-level executable action.
By heuristically integrating the advantages of parameterized skills and the intrinsic capabilities of LLMs, PLAP effectively addresses two key challenges—decision consistency and adaptability. 

We evaluate PLAP in the challenging real-time strategy game environment MicroRTS~\cite{ontanon2018first}, assessing its performance through adversarial matchups across different maps and against multiple baseline opponents. The experiments demonstrate the effectiveness of PLAP, and its best variant outperforms all baseline methods. 
Furthermore, based on Micro-RTS, we develop the Skill-RTS benchmark, which provides comprehensive evaluation metrics to assess the LLMs' skill planning capabilities. 
Building on this benchmark, we evaluate a range of LLMs, both the close-source(GPT, Claude, Gemini) and the open-source(DeepSeek, Qwen). 
The results indicate that GPT-4o achieves the best overall performance, while other models exhibit varying strengths and weaknesses across different evaluation metrics.

Our contributions can be summarized as follows:

\begin{itemize}
    \item We propose PLAP, a two-stage planner-executor planning framework that leverages parameterized skills to effectively ground LLM agents in adversarial long-horizon planning. Include the following key features:
    \begin{itemize}
        \item PLAP operates without extra learning costs and does not rely on expert experience or prior knowledge. 
        \item The dynamically constructed prompt enables the LLM planner to adjust the skill plan according to the evolving situation and opponent's behavior. 
        \item Following the high-level plan, the executor makes real-time decisions with consistency by using predefined skills functions.
    \end{itemize}
    \item We implement PLAP in MicroRTS environment and experimentally validate its effectiveness.
    \item We develop the Skill-RTS benchmark with comprehensive evaluation metrics and establish a leaderboard to assess modern LLMs' skill planning ability.
\end{itemize}

\section{Related Work}
\textbf{LLMs for Long-Horizon Planning.}
Recent work~\cite{ichter2022do, dalal2024planseqlearn, kumar2024practice} has explored leveraging LLMs for long-horizon planning, but most approaches require training auxiliary networks. 
SayCan~\cite{ichter2022do} requires training a value function to filter executable actions; Plan-Seq-Learn~\cite{dalal2024planseqlearn} relies on training a low-level control policy; EES~\cite{kumar2024practice} needs to learn a parameter selection policy to guide execution. 
While effective in robotics tasks, these methods struggle in adversarial settings, where the skill parameter space dynamically changes(e.g. newly trained unseen soldiers). Training an additional parameter-generation strategy in such scenarios is highly challenging, leaving it as a direction for future research.
Other approaches~\cite{bhat2024grounding, nayak2024longhorizon} significantly increase LLM invocation costs. 
LLaMAR~\cite{nayak2024longhorizon} features a more complex architecture, incorporating four LLM-based modules that collaborate to achieve planning and execution.
Though successful in certain domains, these methods become impractical for real-time, decision-intensive tasks due to unaffordable inference time and invocation costs.

\textbf{Parameterized Skills Planning.}
In reinforcement learning, parameterized skills are not a single policy learned for a specific task, but rather a set of skills that can be adapted to a range of related tasks by adjusting certain parameters. 
Some studies~\cite{da2012learning,da2014active,masson2016reinforcement,ames2018learning} focus on parameterized skills learning or planning. 
Unlike previous research on parameterized skill planning, PLAP utilizes predefined parameterized skill functions as a bridge between high-level planning and low-level execution. The design enhances the grounding capabilities of LLM agents in adversarial long-horizon environments while eliminating the need for additional training.

\section{Preliminaries}
\subsection{Parameterized Skills}
Following~\cite{masson2016reinforcement}, we consider a planning task equipped with a library of predefined skills as a parameterized action Markov decision process (PAMDP), which is specifically described by 5-tuple 
$(\mathcal{S},\mathcal{U},R,T,\gamma)$, 
where $\mathcal{S}$ is a set of states, $\mathcal{U}$ is a set of parameterized skills, $R$ is a reward model of environment dynamics, $T$ is the transition model of dynamics, and $\gamma$ is a discount factor.
A parameterized skill is a high-level function that takes multiple continuous or discrete parameter values as the input and ultimately outputs underlying low-level control actions. 
For example, in adversarial domains, \textit{attack(attacker=`Plane1', target=`Enemy1', coordinates=(100,120))} represents a parameterized skill where \textit{attack} is the skill name, and its parameters define the initiator of the attack, the target, and the target's position. 
Formally, each parameterized skill $u_i \in \mathcal{U}$ can be represented by a tuple \cite{sutton1999between}: 
$(\pi_i, \Theta_i, I_i, \beta_i)$, 
where $\pi_i(a \mid s, \theta_i)$ is a low-level control policy that gives the probability of the agent taking action $a$ in state s, given the input parameter $\theta_i$, $\Theta_i \subseteq \mathbb{R}^m$ denotes parameter vector space, 
$I_i: \mathcal{S} \rightarrow \{0,1\}$ characterizes
states where the skill is available, and $\beta_i: \mathcal{S} \rightarrow \{0,1\}$ is the termination condition indicating whether the skill is terminated in state $s$.
In this setting, the agent needs to find a policy mapping the observation state to a distribution over parameterized skill, $(u_i,\theta_i)$, where $u_i$ is a skill and $\theta_i$ is the corresponding parameter vector.

\subsection{Grounding Language Models}
LLMs demonstrate strong instruction-following capabilities~\cite{NEURIPS2020_1457c0d6}, enabling researchers to design task-specific user prompts that guide LLMs to generate responses for a diverse range of general-purpose tasks~\cite{NEURIPS2022_8bb0d291, pmlr-v162-huang22a, ma2023large, tang-etal-2024-medagents, imani-etal-2023-mathprompter, ding2024large}: $\text{response} \sim P(\cdot \mid \text{prompt}, \theta_{LM}),$
where \text{prompt} represents the LLMs input containing task-specific user instructions, \text{response} is the textual output generated by the LLMs, and $\theta_{LM}$ denotes the internal policy parameters.
However, LLMs cannot interact directly with the non-textual environment, which requires environment-executable actions.
To bridge the gap, a post-processing function $f$ is employed to parse the textual response into environment-executable actions $a$ (or parameterized skills $u(\theta)$): $a = f(\text{response}).$


\begin{figure*}[htbp]
    \centering
    \includegraphics[width=\linewidth]{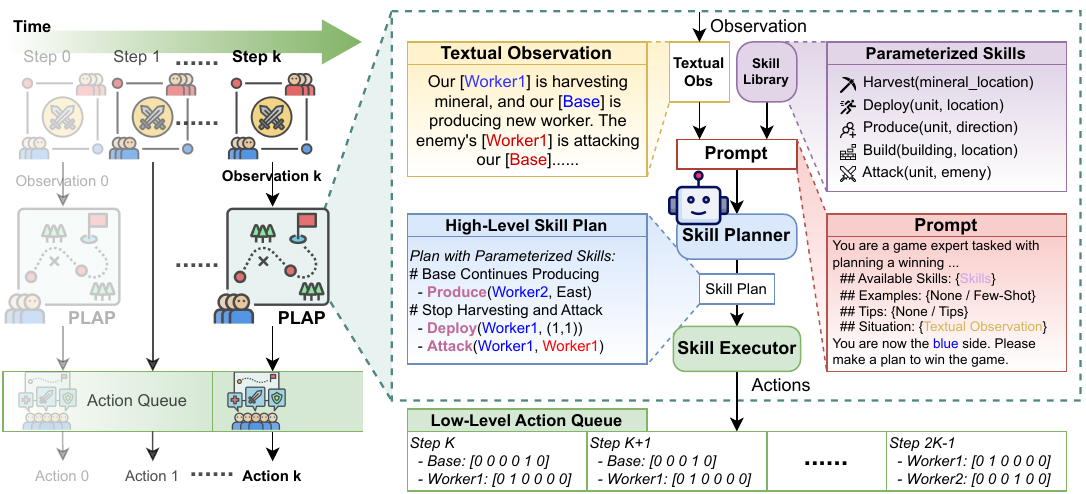}
    \caption{The overall framework of PLAP in the adversarial domain. For every k step, PLAP regenerates an action queue to control further k low-level actions. We demonstrate the detailed generation of PLAP in the right.
    The observation is converted into \colorbox{plap_yellow}{textual observation} and combined with parameterized skill descriptions from the \colorbox{plap_purple}{skill library} as components of the \colorbox{plap_red}{prompt}.
    As a result, the \colorbox{plap_blue}{Skill Planner} generates the high-level skill plan, which is provided to the \colorbox{plap_green}{Skill Executor}, maintaining the low-level action queue during step $[k, 2k-1]$.}
    \label{fig:plap}
\end{figure*}

\section{Approach}
We select the MicroRTS environment~\cite{ontanon2018first, huang2021gym} to investigate the long-horizon planning capabilities of LLMs in adversarial domains.
The red and blue sides compete on symmetrical maps by gathering resources, constructing units, and ultimately defeating the opponent to achieve victory, more details illustrated in Appendix~\ref{apd:urts}. 
In MicroRTS, there are thousands of game steps, each requiring real-time decision-making, which poses a typical long-horizon challenge. 

We introduce the Plan with Language, Act with Parameter (PLAP) framework, a planner-executor structure in which the planner performs skill planning at regular intervals, directing the executor to produce low-level actions for the subsequent game steps.
We set the interval as a hyperparameter $k$ (with k=100 in our experiments).
Specifically, if a game requires $n$ steps to complete, skill planning needs to be performed $n/k$ times, significantly reducing the cost of LLM invocations. 

We illustrate the PLAP's architecture comprising three key components: Skill Planner, Skill Executor, and Skill Library in Fig.~\ref{fig:plap}. Each invocation primarily involves the following steps:
\begin{itemize}
\item \textbf{Prompt Construction}: Construct a skill prompt including the environment’s observations and available skills.
\item \textbf{Plan Generation}: Utilize the skill prompt to query the LLM for generating a skill plan, then verify the validity of the skills and parameters.
\item \textbf{Action Generation}: Input the high-level skill plan into the skill executor to generate a low-level action queue, which is maintained over the subsequent $k$ game steps.
\end{itemize}

\subsection{Skill Prompt}
In this section, we detail the construction of the PLAP prompt, with an overview of the simplified prompt shown in Fig.~\ref{fig:prompt_simp}.
Due to space limitations, verbose or less important content has been omitted from the figure to highlight the key information.
The prompt consists of the following primary components, the user's instruction, the game manual, the available skills, the few-shot examples, the expert tips, and the current battlefield situation.

\begin{figure}[t]
    \centering
    \input{prompt_simp}
    \caption{The illustration of the PLAP prompt structure.}
    \label{fig:prompt_simp}
\end{figure}

\textbf{Concise and Clear Structural.}
The Skill Prompt follows a concise and structural format, with each field clearly marked by `\#\#FIELD'. The `Game Manual' field provides the necessary explanation and details about the environment. 
The structural description of available parameterized skills from the skill library is included in `Available Skills' field. 
The `Battlefield Situation' field includes textual observation of the current game step.

\textbf{Few-Shot PLAP.}
Leveraging the in-context learning capabilities~\cite{NEURIPS2020_1457c0d6} exhibited by LLMs, we also explored the few-shot PLAP method. 
This approach involves providing a few skill planning examples, added to the `Examples’ field, to guide the model in generating more effective plans.
These examples are manually constructed on the basesWorkers8x8 map, with the number limited to 2. The later experimental results demonstrate that few-shot PLAP with just two examples performs effectively. 
For simplicity, we denote zero-shot PLAP and few-shot PLAP as zs-PLAP and fs-PLAP, respectively.

\textbf{PLAP with Expert Tips.}
Some researchers use some purpose-specific tips~\cite{chen2023introspective} from experts acquainted with the environment, aiming to induce better responses from LLM.
The PLAP prompt also contains a `tips' field, which can either be left \textit{None} or filled with several expert tips, such as `\textit{Attack enemy barracks first to prevent them from producing soldiers.}'
Similarly, we denote zero-shot PLAP with expert tips as zs-tip-PLAP and without expert tips still as zs-PLAP. 

Formally, at the $t$ game step, the skill prompt $D_t$ can be expressed as:
\begin{equation}
    D_t \leftarrow \text{Construction}(\text{Inst}, \text{Manu}, \mathcal{U^{\text{desc}}}, \text{shots}, \text{tips}, O_t^{\text{text}}), \label{eq:plap:prompt}
\end{equation}
where “Inst” represents the user’s instruction, “Manu” is the game rules description, $\mathcal{U^{\text{desc}}}$ refers to the description of parameterized skills, and $O_t^{\text{text}}$ represents the real-time textual observation at the $t$ step. 
The carefully designed prompt is then utilized to generate the skill plan. 

\subsection{Skill Plan}
Previous work~\cite{kumar2024practice} utilizes LLM to choose a skill, while the generation of parameters typically relies on prior skill-parameter distributions or additional policy learning. 
In contrast, by leveraging $D_t$, PLAP entrusts both skill planning and parameter selection to the LLM skill planner at the $t$ step.
The following post-processing function $f$ processes the LLM’s response to extract the skill plan. Formally, this process is expressed as follows:
\begin{equation}
    \text{response}_t \sim P(\cdot \mid D_t, \theta_{LM}), \label{eq:llm_gen2} 
\end{equation}
\begin{equation}
u_1(\theta_1),u_2(\theta_2),\dots,u_{c_t}(\theta_{c_t}) = f(\text{response}_t), \label{eq:plap:skill}
\end{equation}
where $c_t$ represents the total number of valid parameterized skills, and $u_i(\theta_i)$ corresponds to the $i$-th parsed skill $u_i$ with its corresponding parameters $\theta_i$. 
For simplicity, we denote the skill plan at $t$ step as $Q_t$: 
\begin{equation}
    Q_t=\{u_1(\theta_1),u_2(\theta_2),\dots,u_{c_t}(\theta_{c_t})\}. \label{eq:plap:skill2}
\end{equation}

\textbf{Post-processing Parser.} The original response consists of multiple lines [SKILL\_NAME](PARAMETERS), enclosed between `START-' and `END OF SKILL\_PLAN'.
The parser $f$ utilizes regular expression matching to extract skills and their parameters.
Any lines that fail to match (e.g., formatting errors, nonexistent skills, or invalid parameters) are discarded. Consequently, the final skill plan consists of $c_t$ valid parameterized skills.

\begin{figure}[t]
    \centering
    \includegraphics[width=1.0\linewidth]{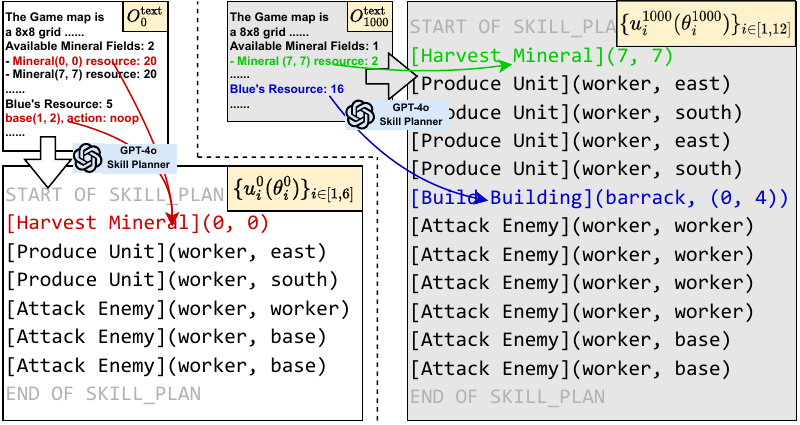}
    \caption{Two examples of skill plans generated by the GPT-4o-driven LLM planner at $t=0$ and \colorbox{skill_plan_1}{$t=1000$}. At $t=0$, there are two mineral fields available at different locations. The planner selects the field closest to the base for resource harvesting(red arrow). 
At t=1000, only one mineral field remains, and the planner gathers resources from this remaining field(green arrow). 
Additionally, with more resources accumulated, the planner plans to construct an advanced building, the barrack(blue arrow).
    }
    \label{fig:skill_plan}
\end{figure}

We show two skill plan examples from GPT-4o-driven skill planner in Fig.~\ref{fig:skill_plan}, where some causal relationships can be observed in the response. 
These examples demonstrate the LLM’s ability to utilize parameterized skills to adaptively plan in response to continuously evolving situations.

\subsection{Skill Execution}
The skill executor is a pivotal component in ensuring the effective execution of the skill plan, as it controls the generation of low-level executable actions.
Since skill planning occurs at intervals of $k$, the skill executor is responsible for producing the action vector $\vec{a_{t'}}$, where $t'\in[t, t+k-1]$, based on skill plan $Q_t$ and current state(observation) $s_{t'}$.
For each skill $u_i = (\pi_i, \Theta_i, I_i, \beta_i)$ in $Q_t$, the executor first determines whether the skill is active in $s_{t'}$ using $I_i$. Active skills are grouped into the set $Q_{t'}^{\text{active}}$, defined as: 
\begin{equation}
    Q_{t'}^{\text{active}}=\{u_i \in Q_t \mid I_i(s_{t'})=1\}. \label{eq:plap:exe1}
\end{equation}
Then for $u_i \in Q_{t'}^{\text{active}}$, the corresponding policy $\pi_i$, a pre-implemented function, is used to generate a sequence of actions to perform the skill, given the current state(observation) $s_{t'}$ and parameter $\theta_i$: 
\begin{equation}
    a_{(t',0)}^i, a_{(t',1)}^i, a_{(t',2)}^i, \dots=\pi_i(s_{t'}, \theta_i), \forall u_i \in Q_{t'}^{\text{active}}. \label{eq:plap:exe2}
\end{equation}
The skill executor collects the first action $a_{(t',0)}^i$ generated by $\pi_i$ as $t'$-step action of skill $u_i$. These individual control actions are then aggregated into the action vector, defined as: 
\begin{equation}
    \vec{a_{t'}}=[a_{(t',0)}^0, a_{(t',0)}^1, \cdots, a_{(t',0)}^{c_{t'}}], \label{eq:plap:exe3}
\end{equation}
where $c_{t'}$ represents the size of $Q_{t'}^{\text{active}}$.

It is important to highlight that, at the next step $t'+1$, the executor does not directly use the second action $a_{(t',1)}^i$ from the previously generated sequence to form the next action vector $\vec{a_{t'+1}}$. This decision is based on two key observations:
1. The skill $u_i$ may become inactive at state $s_{t'+1}$ due to changes in the environment.
2. The action $a_{(t',1)}^i$ may no longer be valid or appropriate in the new state, particularly due to the unpredictable behaviors of the opponent.

Therefore, the executor uses the policy $\pi_i$ to generate a new action sequence based on state $s_{t'+1}$ for each active skill $u_i \in Q_{t'+1}^{\text{active}}$. The action vector $\vec{a_{t'+1}}$ is then constructed by aggregating the newly generated first actions:
\[
\vec{a_{t'+1}} = [a_{(t'+1,0)}^0, a_{(t'+1,0)}^1, \dots, a_{(t'+1,0)}^{c_{t'+1}}].
\]
This iterative process ensures that the actions executed remain relevant and effective, dynamically adapting to changes in the environment and opponent behavior.

\textbf{Termination Condition.} The termination condition $\beta_i$ is hard-coded in the corresponding skill function. Once the termination condition of $u_i$ holds at state $s_{t''}$, where $t''\in[t, t+k-1]$, $u_i$ is deleted from the skill plan $Q_t$: 
\begin{equation}
    Q_{t} \leftarrow Q_{t} \setminus \{u_i \in Q_t\mid \beta_i(s_{t''})=1 \},
    \label{eq:plap:exe4:delete}
\end{equation}
which enables the executor to maintain the $Q_t$ dynamically. 

In summary, the skill executor controls the generation of low-level action vector $\vec{a_{t'}}$, where $t'\in[t,t+k-1]$. It ensures adherence to the skill plan $Q_t$ produced by the LLM-based skill planner, which operates every $k$-step interval. 
The skill library contains all available skills and associated functions, supporting the skill planner in plan generation and the executor in action generation.
These components form PLAP, a cohesive system for high-level planning and low-level execution.
The formal representation of PLAP is summarized in Algorithm~\ref{alg:plap}.

\begin{algorithm}[h]
\caption{\textbf{PLAP}: \textbf{P}lan with \textbf{L}anguage, \textbf{A}ct with \textbf{P}arameter}
\label{alg:plap}
\begin{algorithmic}[1]
    \REQUIRE The environment dynamics $\text{Env}$; the game step $t$; the state(observation) $s_t$; the planning interval $k$; the skill prompt $D_t$; the skill plan $Q$; the active skills $Q_t^{\text{active}}$ at state $s_t$; the low-level action $\vec{a_t}$; the set of parameterized skills $\mathcal{U}=\{u_i(\pi_i, \Theta_i, I_i, \beta_i)\}$
    \ENSURE 
    \STATE $t \leftarrow 0$
    \STATE $s_t \leftarrow$ Env.reset()
    \STATE $Q \leftarrow \varnothing$, init the skill plan
    \WHILE {$s_t$ is not terminated state}
        \IF{$t\%k==0$} 
            \STATE \textit{High-level Skill Plan}
            \STATE $D_t \leftarrow $ prompt construction, in \eqref{eq:plap:prompt}
            \STATE $Q \leftarrow \text{Planner}(\text{LLM}, D_t)$, in \eqref{eq:llm_gen2}, \eqref{eq:plap:skill}, \eqref{eq:plap:skill2}
        \ENDIF
        \STATE \textit{Low-level Actions}
        \STATE Compute active skills $Q_t^{\text{active}}$ in \eqref{eq:plap:exe1}
        \STATE $\vec{a_t} \leftarrow \text{Executor}(Q_t^{\text{active}},s_t)$, in \eqref{eq:plap:exe2}, \eqref{eq:plap:exe3}
        \STATE $s' \leftarrow \text{Env.step}(s_t,\vec{a_t})$
        \STATE Update $Q$ based on \eqref{eq:plap:exe4:delete}
        \STATE $t \leftarrow t+1$
        \STATE $s_t \leftarrow s'$
    \ENDWHILE
\end{algorithmic}
\end{algorithm}


\section{Experiments}
\subsection{Skill-RTS Benchmark}
\begin{figure}[t]
    \centering
    \includegraphics[width=1.0\linewidth]{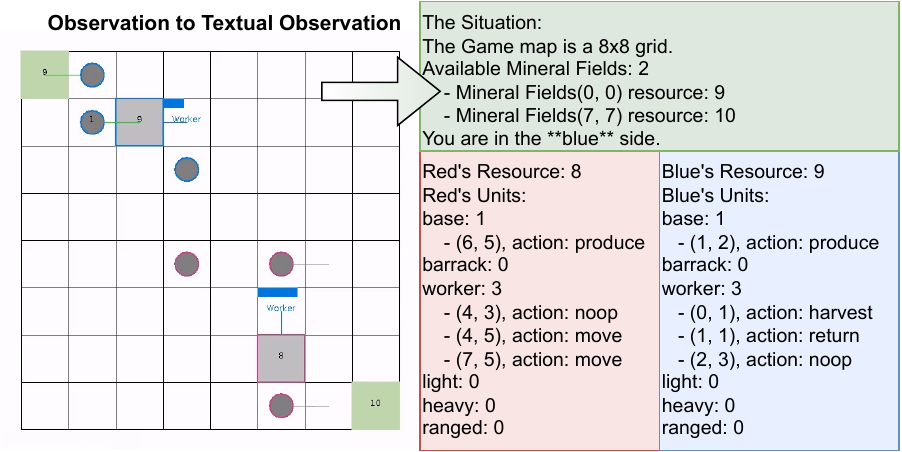}
    \caption{An example of converting original observation into textual observation.}
    \label{fig:obs2text}
\end{figure}

Skill-RTS, a long-horizon skill planning environment, is based on MicroRTS~\cite{huang2021gym, ontanon2013combinatorial}. 
The original observation space and action space of MicroRTS are detailed in Appendix~\ref{apd:urts}.
We transform the original tensor-based observations into a structured description in natural language, illustrated in Fig.~\ref{fig:obs2text}.
We mainly encapsulate its atomic actions into five commonly used parameterized skills, with each parameterized skill potentially containing one or more underlying actions.
The LLM planner only needs to output the skill name and its corresponding parameters. 
Table~\ref{tab:skill_space} lists 5 carefully designed, intuitive LLM-friendly parameterized skill functions. By leveraging these skills, LLMs can efficiently plan game strategies. 

\begin{table*}[t]
    \centering
    \caption{The Skill Space in Skill-RTS}
    \centerline{
    \resizebox{1.0\linewidth}{!}{
    \begin{tabular}{@{}cccp{0.35\linewidth}@{}}
        \toprule
         \textbf{Skill}&  \textbf{Parameters}& \textbf{Atomic Actions Included} &\textbf{Description}\\
         \midrule
         Deploy Unit&  unit type, deployment location&  move, NOOP&Moves a unit of a specified type to a specified location.\\
         Harvest Mineral&  mineral location&  move, harvest, return&Move to a specified location to harvest resources and return to base.\\
         Build Building&  building type, building location&  move, produce&Build a building of the specified type at the specified location.\\
         Produce Unit&  unit type, produce direction&  produce&Produce a specified type of unit from a specified direction.\\
         Attack Enemy&  unit type, enemy type&  move, Attack&Command a specified type of unit to attack a specified enemy type of unit.\\
         \bottomrule
    \end{tabular}\label{tab:skill_space}
    }}
\end{table*}

\textbf{Evaluation.} 
To assess the performance of PLAP and highlight the planning capabilities of different LLM-driven skill planners, we establish comprehensive evaluation metrics:
\begin{itemize}
    \item \textbf{Scores}: By competing in multiple matches against five fixed baseline agents, the average score is calculated, assigning +1 for a win, 0 for a draw, and -1 for a loss.
    \item \textbf{Win Rate(WR)}: By calculating the win rate against different LLM-driven PLAP agents, we evaluate the overall skill planning capability of the LLM.
    \item \textbf{Resource Harvesting Ratio(RHR)}: Reflects the efficiency of gathering economic resources.
    
    $\qquad \text{RHR} = \text{resources\_harvested} / \text{game\_time} * 100$

    \item \textbf{Resource Utilization Rate(RUR)}: Evaluates the efficiency of resource use, reflecting the economic management effectiveness.
    
    $\qquad \text{RUR} = {\text{resources\_spent}} / {\text{game\_time}} * 100$

    \item \textbf{Unit Production Rate(UPR)}: Indicates how efficient the model is at producing units.
    
    $\qquad \text{UPR} = {\text{unit\_production}} / {\text{game\_time}} * 100$
    
    \item \textbf{Combat Efficiency Ratio(CER)}: Measures the offensive power by calculating the ratio of damage dealt to damage taken during battles. 
    
    $\qquad \text{CER} = \text{damage\_dealt} / \text{damage\_taken}$
\end{itemize}

\subsection{Experiment Setting}
We test experimentally both open-source LLMs and commercial LLMs. Since our method does not rely on training or fine-tuning, the mainly considered hyper-parameters are temperature and max\_tokens and we set temperature to 0, max\_tokens to 256 tokens unless otherwise specified. The default planning interval, $k$, is set to 100 steps in the experiment.
The game map mainly used to evaluate the performance is `basesWorkers8x8', which is a standard and balanced map starting with one base and one worker respectively. 
Since the construction of LLMs' prompt is independent of the specific map, more experimentally assessment on various maps can be conveniently implemented based on our framework. 
All LLMs that appeared in our experiments are included in \{Qwen2-72B-Instruct, DeepSeek V2.5, Gemini 1.5 Flash 002, Claude 3 Haiku, Claude 3.5 Sonnet, GPT-3.5 Turbo, GPT-4o mini, and GPT-4o\}\footnote{The detail versions are: claude-3-haiku-20240307, claude-3-5-sonnet-20240620, gpt-3.5-turbo-0125, gpt-4o-mini-2024-07-18, gpt-4o-2024-08-06.}. For simplicity and convenience, model names may be abbreviated in legends or analyses, provided there is no ambiguity.

\subsection{Skill-RTS, an unfamiliar scene for most LLMs}
\begin{figure}[t]
    \centering
    \includegraphics[width=1.0\linewidth]{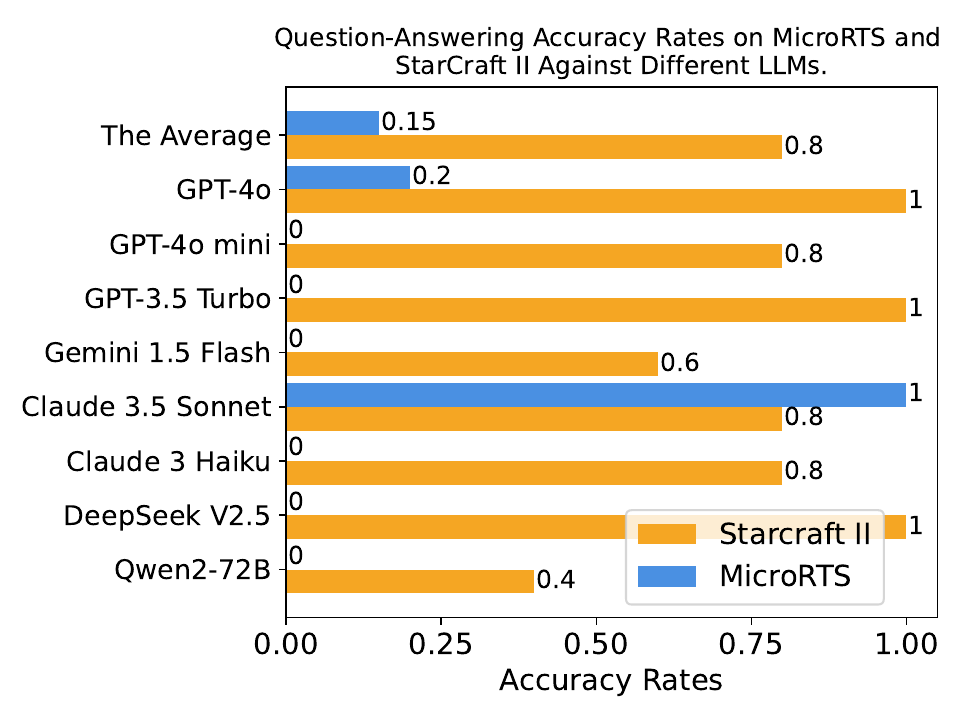}
    \caption{The QA accuracy, assessing LLMs' degree of familiarity.}
    \label{fig:exp0}
\end{figure}

We first show that most LLMs are unfamiliar with Skill-RTS, specifically, we create a set of straightforward question-answer pairs related to the MicroRTS environment and measure their accuracy to assess the degree of familiarity.
An additional question-answer set related to StarCraft II~\cite{vinyals2017starcraft, vinyals2019grandmaster} is designed to serve as a comparative group. StarCraft II exists domain knowledge leakage, as relevant knowledge and expert planning examples have already been learned (or memorized) from pretraining corpora~\cite{ma2023large}.
The result in Fig.~\ref{fig:exp0} reveals that the average accuracy rate of eight various LLMs on the QA is 0.15\%, as a comparison, that on StarCraft II is up to 80\%. 
Notably, only two LLMs achieve non-zero QA accuracy rates on MicroRTS, with the Claude 3.5 Sonnet model demonstrating a 100\% accuracy rate. This suggests that the Sonnet's training data includes the source of questions.
Overall, our findings support the hypothesis that \textbf{most} existing LLMs have little internal knowledge about Skill-RTS, compared to popular StarCraft II. 
For transparency, we include all questions used to assess domain-related knowledge of LLMs, along with the construction process, in Appendix~\ref{apd:qa}.

\subsection{PLAP, an outstanding grounding solution}
\begin{figure}[t]
    \centering
    \includegraphics[width=1.0\linewidth]{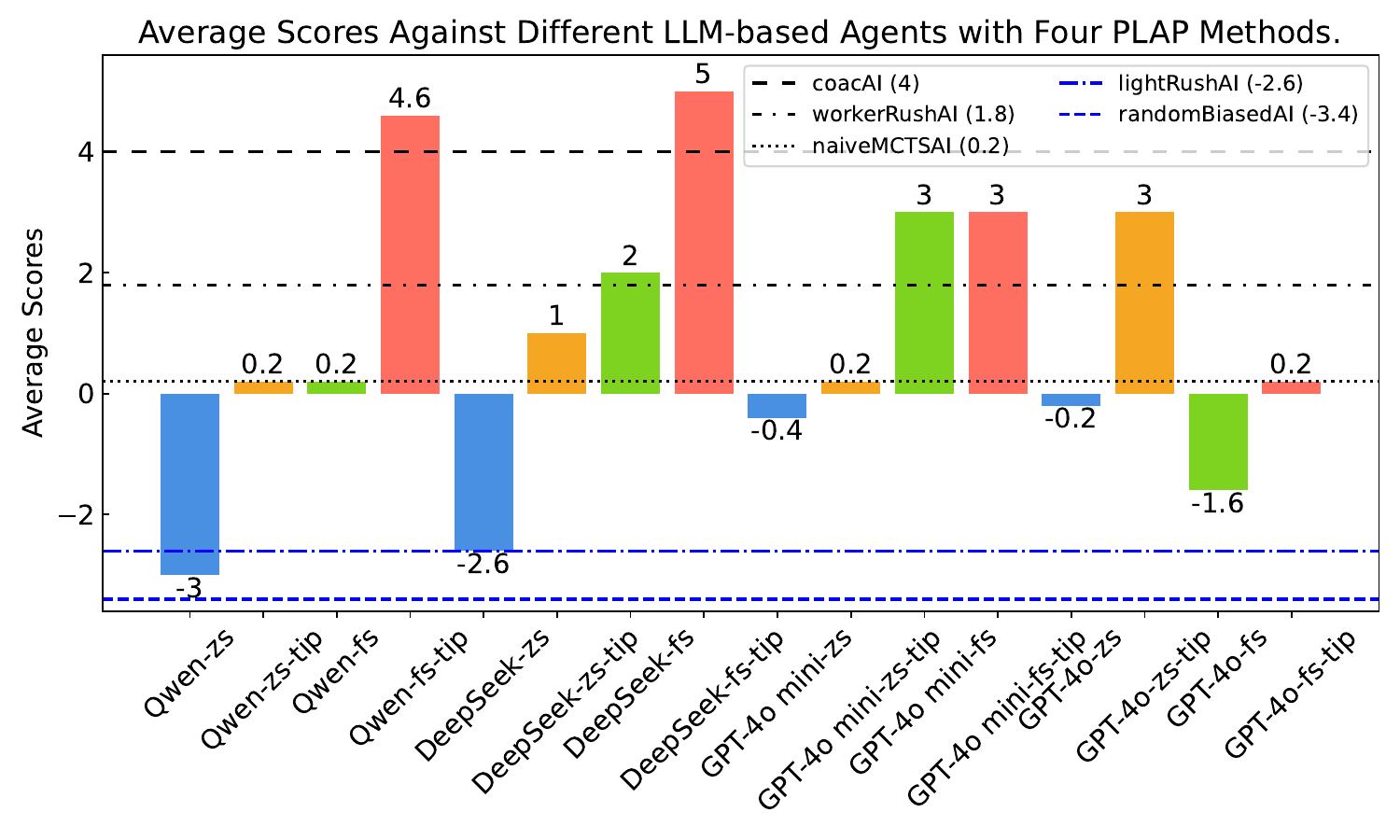}
    \caption{The illustration shows the average scores of four PLAP variants: \colorbox{plap_blue}{zero-shot}, \colorbox{plap_yellow}{zero-shot with tips}, \colorbox{plap_green}{few-shot}, and \colorbox{plap_red}{few-shot with tips} PLAP. These scores are compared to five baseline AI bots, with their levels represented by different dashed lines.}
    \label{fig:llms_bots}
\end{figure}

To comprehensively evaluate the performance of the PLAP framework, we implemented four variations based on the PLAP framework, respectively zero-shot PLAP, zero-shot PLAP with tips, few-shot PLAP, and few-shot PLAP with tips. 
Since the PLAP framework is independent of reasoning modular, it can be easily extended to incorporate additional methods, such as PLAP with Chain-of-Thought~\cite{wei2022chain}, Tree-of-Thought~\cite{yao2024tree}, Reflection~\cite{shinn2024reflexion}, and React~\cite{yao2023react} reasoning.
We assess four LLM agents' effectiveness by battling with bots of different levels, including RandomBiasedAI, LightRush, NaiveMCTS, WorkerRush, and CoacAI, winner of the 2020 competition~\footnote{The 2020 COG Results can be found at: \url{https://sites.google.com/site/micrortsaicompetition/competition-results/2020-cog-results}}. 
A detailed description of the five baselines is provided in Appendix~\ref{apd:robot}.
We evaluate the performance based on \textbf{Scores}. For each model and each prompting method, five rounds will be played against the baseline rule-based AI.
We illustrate the results in Fig.~\ref{fig:llms_bots}.
There are \textbf{three main findings} as follows.

\subsubsection{PLAP Methods Surpass Most Baselines}
The average score of all 16 settings is about 0.91, which is between the WorkerRush(1.8) and naiveMCTS(0.2) algorithms, therefore, we can conclude that PLAP is completely superior to RandomBiasedAI, LightRush, and naiveMCTS. 
Some methods like Deepseek-fs, GPT-4omini-fs, GPT-4o-mini-fs-tip, and GPT-4o-zs-tip can outperform WorkerRush. Meanwhile, Qwen-fs-tip and DeepSeek-fs-tip demonstrate superior results compared to the best bot, CoacAI. 

\subsubsection{Few-Shot PLAP Methods \textbf{NOT} Always Outperform the Zero-Shot}
Fig.~\ref{fig:llms_bots} shows an obvious tendency in the increasing scores between zero-shot PLAP and few-shot PLAP methods. Specifically, using two carefully designed examples has an average improvement of about 3.8 scores for the former three LLMs.
However, this rule does not hold for GPT-4o model, which shows a reduction of 1.4 scores for GPT-4o-fs and 2.8 scores for GPT-4o-fs-tip compared to zero-shot prompting.
A reasonable explanation for this phenomenon is that the GPT-4o is a more advanced model, surpassing the performance level of the other three LLMs.
Clear and comprehensive instructions in the prompt are sufficient to activate its task-solving abilities, and adding examples may negatively impact performance.

\subsubsection{Expert Tips are Remedy}
From the visualized results, it can be observed that the addition of expert tips leads to consistent improvements in both zero-shot and few-shot methods. That suggests that expert tips can efficiently enhance the performance of PLAP.

\subsection{LLM Leaderboard on Skill-RTS Bechmark}
\begin{figure}[t]
    \centering
    \includegraphics[width=1.0\linewidth]{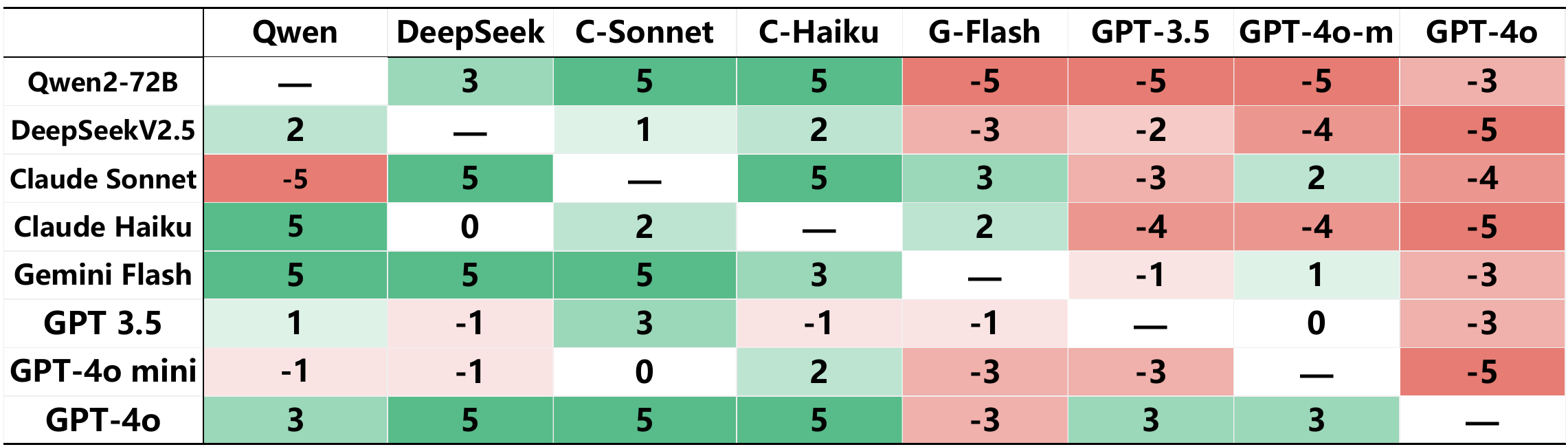}
    \caption{The matchup details of the leaderboard. Each row represents a model playing as the Blue, while each column represents a model playing as the Red. In each \textbf{Scores} cell, \colorbox{plap_green}{higher values} indicate a higher score for the Blue side, whereas \colorbox{plap_red}{lower values} indicate a higher score for the Red side. The score range is [-5, 5].}
    \label{fig:llmvsllm_detail}
\end{figure}

To furthermore reveal the LLM agents' grounding ability based on parameterized skill planning, we designed a series of Red-Blue Confrontations between different LLM agents employing the same \textbf{zero-shot PLAP} method. 
The candidate models consist of Qwen2-72B-Instruct, DeepSeek V2.5, Gemini 1.5 Flash, Claude 3 Haiku, Claude 3.5 Sonnet, GPT-3.5 Turbo, GPT-4o mini, and GPT-4o. 
Each LLM serves as both the Red and Blue side, competing against the other seven models. Each confrontation consists of five matches, with scores ranging from -5 to 5. In total, each model participates in 70 matches. The detail of matches is shown in Fig.~\ref{fig:llmvsllm_detail}.

As illustrated in Table~\ref{tab:leaderboard}, we established a leaderboard according to the results of the confrontation. Rankings are determined based on \textbf{Scores}, with multiple performance metrics calculated from the confrontation results. Bold values indicate the best performance, while underlined values represent the two next-best results.
We observe that the LLM agents powered by GPT-4o and Gemini 1.5 Flash achieve win rates of 84.3\% and 65.7\%, respectively, securing first and second place in Skill-RTS. 
Additionally, different models exhibit distinct skill preferences. For example, GPT-3.5 Turbo excels at spending resources efficiently and rapidly constructing units, while Qwen2-72B demonstrates stronger capabilities in resource gathering and combat.

\begin{table}[t]
    \centering
    \caption{The LLM leaderboard on Skill-RTS.}
    \begin{tabular}{@{}lccccccc@{}} 
        \toprule
        \textbf{Model} & \textbf{Scores} & \textbf{WR} & \textbf{RHR} & \textbf{RUR} & \textbf{UPR} & \textbf{CER} \\
        \midrule
        GPT-4o & \textbf{49} & \textbf{84.3\%} & \textbf{2.32} & \textbf{2.34} & \textbf{1.61} & \textbf{1.63} \\
        Gemini 1.5 Flash & \underline{25} & \underline{65.7\%} & \underline{1.93} & \underline{1.57} & \underline{1.57} & \underline{1.22} \\
        GPT-3.5 Turbo & \underline{13} & \underline{52.9\%} & 1.13 & \underline{1.38} & \underline{1.32} & 0.95 \\
        GPT-4o mini & -4 & 40.0\% & 1.08 & 1.27 & 1.22 & 0.81 \\
        Qwen2-72B-Instruct & -15 & 35.7\% & \underline{1.61} & 1.38 & 0.90 & \underline{1.26} \\
        Claude 3.5 Sonnet & -18 & 28.6\% & 1.56 & 1.35 & 1.29 & 1.14 \\
        Claude 3 Haiku & -25 & 20.0\% & 0.98 & 1.23 & 1.20 & 0.89 \\
        DeepSeek V2.5 & -25 & 18.6\% & 0.98 & 1.09 & 0.83 & 0.70 \\
        \bottomrule
    \end{tabular}
    \label{tab:leaderboard}
\end{table}

\section{Conclusion}
In this paper, we propose the PLAP framework, which utilizes a set of predefined parameterized skills to ground the LLM agents in adversarial long-horizon environments. 
The planner continuously senses environmental and opponent changes by in-context learning, generating an adaptive skill plan. The executor adheres to the plan by invoking manually designed skill functions, ensuring consistent low-level control actions.
PLAP requires no additional training, and by generating plans every $k$ steps, PLAP reduces the calling frequency, improving the efficiency in decision-intensive tasks. 
We validate PLAP through experiments, demonstrating that zero-shot PLAP, powered by a GPT-4o planner, outperforms 80\% of baseline AI bots. Moreover, by incorporating well-crafted examples and expert tips, Qwen2-72B surpasses all five baselines.

To advance the development of this field, we developed the Skill-RTS benchmark, featuring diverse evaluation metrics to assess LLM agents’ parameterized skill planning ability. 
Additionally, we release a skill planning leaderboard, ranking modern LLMs based on their performance on the Skill-RTS benchmark.
GPT-4o and Gemini 1.5 Flash emerge as the top-performing models on the leaderboard. Meanwhile, some other models exhibit strengths in different aspects. 

We discuss the several limitations of our study and directions for future research:
\begin{itemize}
    \item Limited Evaluation Scope: PLAP has only been tested in the MicroRTS environment, lacking validation in more realistic adversarial scenarios. Future research will extend evaluations to complex, real-world domains.
    \item Parameterized Skills Design: The skill functions used in PLAP are manually constructed, which limits generalization across different domains. A potential future direction is to explore LLM-based code generation, reducing human effort and enhancing generalization.
    \item Opponent Strategy Modeling: PLAP currently focuses on explicit state(observation) changes in the environment and opponent, without modeling the opponent’s strategic intent. This limits its upper performance bound. Future work could consider modeling and exploiting opponents to enhance decision-making.
\end{itemize}

Despite these limitations, our proposed PLAP provides an initial exploration of integrating LLMs with parameterized skills to address the challenges of adversarial long-horizon tasks.
We also construct the Skill-RTS benchmark, a valuable platform for advancing this research area.
Ultimately, this work lays the foundation for applying LLM-driven skill planning in real-world adversarial scenarios.

\section{Appendix}
\subsection{MicroRTS Environment} \label{apd:urts}

\begin{figure}[t]
    \centering
    \includegraphics[width=0.9\linewidth]{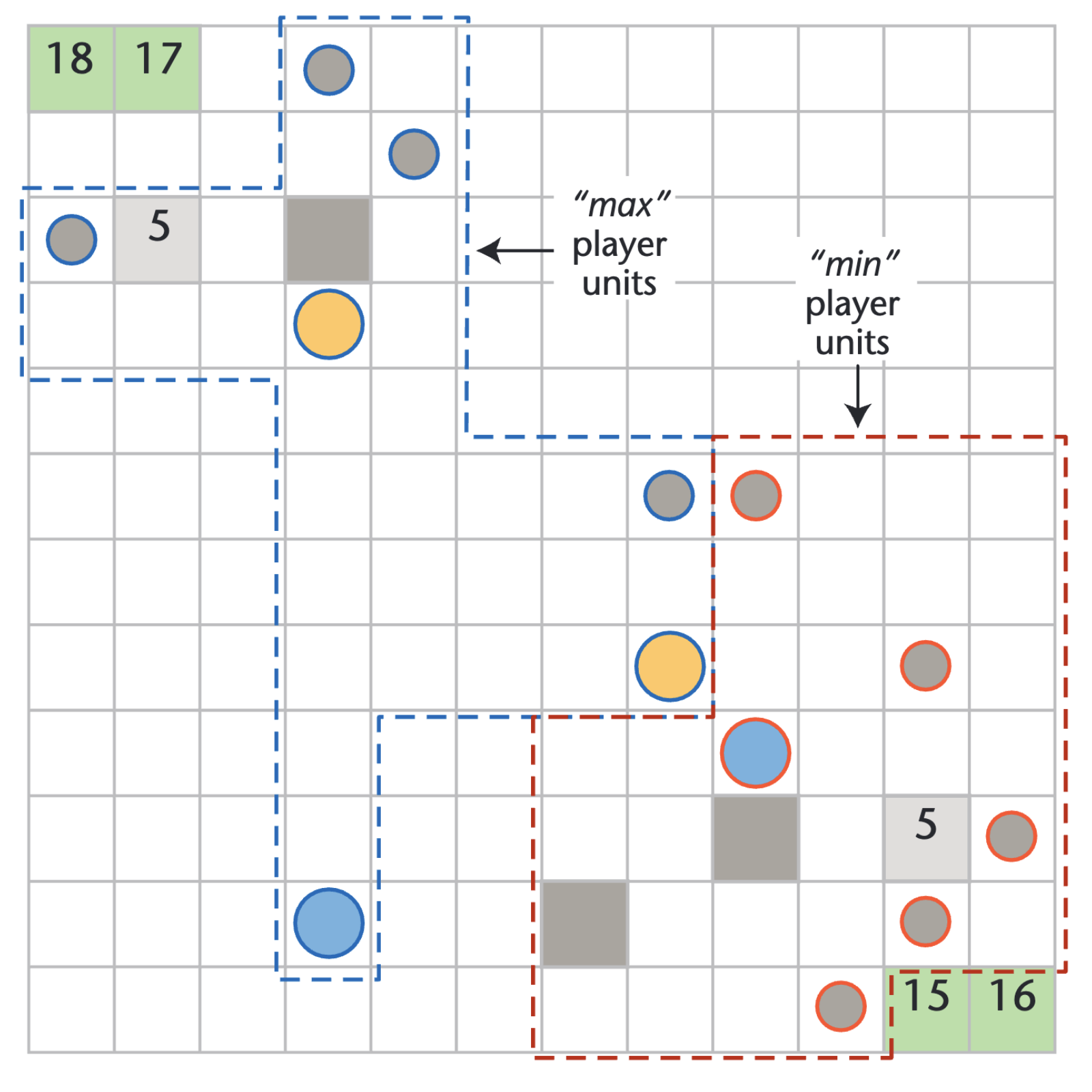}
    \caption{\textbf{A Screenshot~\cite{ontanon2018first} of the MicroRTS Environment.} The “max”/“min” player units refer to the units belonging to the blue/red side. Square-shaped units represent different structures: bases (light grey) for generating workers, barracks (dark grey) for creating military forces, and resource mines (green) where workers gather materials to enable further unit production. Circular units signify entities, with smaller dark grey circles denoting workers and larger yellow or light blue circles indicating military units. }
    \label{fig:urts-screenshot}
\end{figure}

To provide a more intuitive explanation of the MicroRTS environment, we reference the example figure from the paper~\cite{ontanon2018first}, shown in Fig.~\ref{fig:urts-screenshot}.

In a map of size \(h \times w\), the observation returned by the MicroRTS environment is a tensor with a shape of \((h, w, n_f)\), where \(n_f\) represents the number of feature planes, which has a value of 27, as shown in Table~\ref{tab:observation_features}. These different feature planes are obtained by concatenating multiple one-hot encoded features. 

\begin{table}[t]
    \centering
    \caption{Observation Features in MicroRTS}
    \centerline{
    \resizebox{0.9\linewidth}{!}{
    \begin{tabular}{@{}lp{0.1\linewidth}p{0.5\linewidth}@{}}
        \toprule
        \textbf{Features} & \textbf{Planes} & \textbf{Description} \\ 
        \midrule
        Hit Points       & 5 & 0, 1, 2, 3, $\geq$ 4 \\
        Resources        & 5 & 0, 1, 2, 3, $\geq$ 4 \\
        Owner            & 3 & player 1, -, player 2 \\
        Unit Types       & 8 & -, resource, base, barrack, worker, light, heavy, ranged \\
        Current Action   & 6 & -, move, harvest, return, produce, attack \\
        \bottomrule
    \end{tabular}
    \label{tab:observation_features}
    }}
\end{table}

For action space, shown in Table~\ref{tab:urts_action_space}, each action has a $78$ length of action encoding in MicroRTS.
Thus, there is a $78 \times wh$-dimensional action vector in the $w\times h$ grids map.
For example, if the agent issues a command to a worker at position (0,0) to build a barrack to the south in a $2\times 2$ map, the action will be encoded in the following way:
\begin{align*}
    \mathbf{a}_1 &= [4, 0, 0, 0, 2, 2, 0] \\
    \mathbf{a}_2 &= [0, 0, 0, 0, 0, 0, 0] \\
    \mathbf{a}_3 &= [0, 0, 0, 0, 0, 0, 0] \\
    \mathbf{a}_4 &= [0, 0, 0, 0, 0, 0, 0] \\
    \text{action\_tensor} &= \text{one\_hot}(\text{concat}(\mathbf{a}_1, \mathbf{a}_2, \mathbf{a}_3, \mathbf{a}_4))
\end{align*}

\begin{table}[t]
    \centering
    \caption{The Atomic Action Components in MicroRTS. $a_r=7$ in the maximum attack range.}
    \centerline{
    \resizebox{1.0\linewidth}{!}{
    \begin{tabular}{cccc}
    \toprule
     \textbf{Action Components}& \textbf{Range}& \multicolumn{2}{c}{\textbf{Description}}\\
     \midrule
     Action Type& $[0,5]$& \multicolumn{2}{l}{NOOP, move, harvest, return, produce, attack}\\
     Move Parameter& $[0,3]$& \multicolumn{2}{l}{north, east, south, west}\\
     Harvest Parameter& $[0,3]$& \multicolumn{2}{l}{north, east, south, west}\\
     Return Parameter	
    & $[0,3]$& \multicolumn{2}{l}{north, east, south, west}\\
     Produce Direction Parameter	
    & $[0,3]$& \multicolumn{2}{l}{north, east, south, west}\\
     Produce Type Parameter	
    & $[0,6]$& \multicolumn{2}{l}{resource, base, barrack, worker, light, heavy, ranged}\\
     Relative Attack Position	
    & $[0,a_r^2-1]$& \multicolumn{2}{l}{the relative location of the unit that will be attacked}\\
    \bottomrule
    \end{tabular}\label{tab:urts_action_space}
    }}
\end{table}

\subsection{Question-Answering on MicroRTS} \label{apd:qa}
The question-answer pairs consist of five straightforward questions manually designed by authors, shown below.

{\footnotesize
\begin{enumerate}
\item {How many time units does it take to build the Base in MicroRTS?}
\item {How many hit points of the Barrack in MicroRTS?}
\item How many resources does it cost to build the Light in MicroRTS?
\item How many time units does it take to build the Light in MicroRTS?
\item How much damage does the attack of the Ranged in MicroRTS?
\end{enumerate}}
{\footnotesize
\begin{enumerate}
\item How many hit points of Terran SCV in StarCraft II (SC2)?
\item How many hit points of Zerg Viper in StarCraft II (SC2)?
\item How many hit points of Terran Thor in StarCraft II (SC2)?
\item How many transport slots of Medivac dropship in StarCraft II (SC2)?
\item How many minerals does it cost to produce a Stalker of Protoss in StarCraft II (SC2)?
\end{enumerate}}

These questions are extracted simply from the corresponding wiki of MicroRTS\footnote{The Wiki URL: \url{https://github.com/Farama-Foundation/MicroRTS/wiki}} and StarCraft II\footnote{The StarCraft II Wiki URL: \url{https://starcraft.fandom.com/wiki/StarCraft_II}}, thus there no need extra reasoning to answer the questions. The accuracy of answering can indicate the internal environment-related knowledge of LLMs to a certain degree. 


\subsection{Baseline Bots} \label{apd:robot}
We have chosen three bots with hard-coded strategies, one search-based bot, and the CoacAI as the best baseline, detailed descriptions\footnote{The source is: \url{https://github.com/Farama-Foundation/MicroRTS/wiki/Artificial-Intelligence}} as follows:
\begin{enumerate}
    \item RandomBiasedAI: The RandomBiasedAI moves at random, but with a strong bias towards attacking, harvesting, and returning (5 times more probability).
    \item LightRush: The LightRushAI implements a rush strategy that consists of: training one worker and making it gather resources. Once there are enough resources for building barracks, build a barracks. From that point on, train Light units constantly, and send them to attack immediately to the nearest enemy unit.
    \item WorkerRush: The WorkerRush AI implements another rush strategy that consists of: training workers continuously, making one of them gather resources, and sending all the other ones to attack immediately to the closest enemy unit.
    \item NaiveMCTS: The NaiveMCTSAI is a new Monte-Carlo Tree Search algorithm that combines the naive-sampling idea with MCTS. 
    \item CoacAI: The CoacAI is the winner of the 2020 competition, designed with handcrafted rules by human experts.
\end{enumerate}

\section*{Acknowledgment}
We thank our supervisor and mentor for their guidance and valuable advice for this work. 
All authors declare that no conflicts of interest.

\bibliographystyle{IEEEtran} 
\bibliography{ijcnn}

\end{document}